\documentclass[journal,compsoc]{IEEEtran}

\usepackage{spconf,amsmath,graphicx}

\usepackage{amssymb}
\usepackage{cite}
\usepackage{url}
\usepackage{hyperref}
\usepackage[numbered]{bookmark}
\usepackage{lipsum}

\title{TRAINING PROBABILISTIC SPIKING NEURAL NETWORKS WITH FIRST-TO-SPIKE DECODING}

\name{Alireza Bagheri$^\dag$, Osvaldo Simeone$^\ddag$, and Bipin Rajendran$^\dag$}
\address{$^\dag$Department of Electrical and Computer Engineering (ECE),\\ New Jersey Institute of Technology, Newark, NJ 07102, USA.\\
Email: $\{ab745; bipin\}@njit.edu$\\
$^\ddag$Centre for Telecommunications Research, Department of Informatics, \\King's College London, London, WC2R 2LS, UK.\\
Email: $osvaldo.simeone@kcl.ac.uk$}
\begin{document}
\setlength{\abovedisplayskip}{.05pt}
\setlength{\belowdisplayskip}{.05pt}
\maketitle
\begin{abstract}
Third-generation neural networks, or Spiking Neural Networks (SNNs), aim at harnessing the energy efficiency of spike-domain processing by building on computing elements that operate on, and exchange, spikes. In this paper, the problem of training a two-layer SNN is studied for the purpose of classification, under a Generalized Linear Model (GLM) probabilistic neural model that was previously considered within the computational neuroscience literature. Conventional classification rules for SNNs operate offline based on the number of output spikes at each output neuron. In contrast, a novel training method is proposed here for a first-to-spike decoding rule, whereby the SNN can perform an early classification decision once spike firing is detected at an output neuron. Numerical results bring insights into the optimal parameter selection for the GLM neuron and on the accuracy-complexity trade-off performance of conventional and first-to-spike decoding.
\end{abstract}
%
\section{INTRODUCTION}\label{sec:Intro}
%

Most current machine learning methods rely on second-generation neural networks, which consist of simple static non-linear neurons. In contrast, neurons in the human brain are known to communicate by means of sparse spiking processes. As a result, they are mostly inactive, and energy is consumed sporadically. Third-generation neural networks, or Spiking Neural Networks (SNNs), aim at harnessing the energy efficiency of spike-domain processing by building on computing elements that operate on, and exchange, spikes \cite{paugam2012computing}. SNNs can be natively implemented on neuromorphic chips that are currently being developed within academic projects and by major chip manufacturers. Proof-of-concept implementations have shown remarkable energy savings by multiple orders of magnitude with respect to second-generation neural networks (see, e.g., \cite{Inte_web, diamond2016comparing}).

Notwithstanding the potential of SNNs, a significant stumbling block to their adoption is the dearth of flexible and effective learning algorithms. Most existing algorithms are based on variations of the unsupervised mechanism of Spike-Timing Dependent Plasticity (STDP), which updates synaptic weights based on local input and output spikes, and supervised variations that leverage global feedback \cite{ponulak2010supervised, florian2007reinforcement}. Another common approach is to convert trained second-generation networks to SNNs \cite{o2016deep, hunsberger2015spiking}. Among the learning methods that attempt to directly maximize a spike-domain performance criterion, most techniques assume deterministic Spike Response Model (SRM) neurons, and propose various approximations to cope with the non-differentiability of the neurons' outputs (see \cite{anwani2015normad, lee2016training} and references therein). 
%
\begin{figure}[t]
	\begin{minipage}[b]{1.0\linewidth}
		\centering
		\centerline{\includegraphics[scale = 0.5]{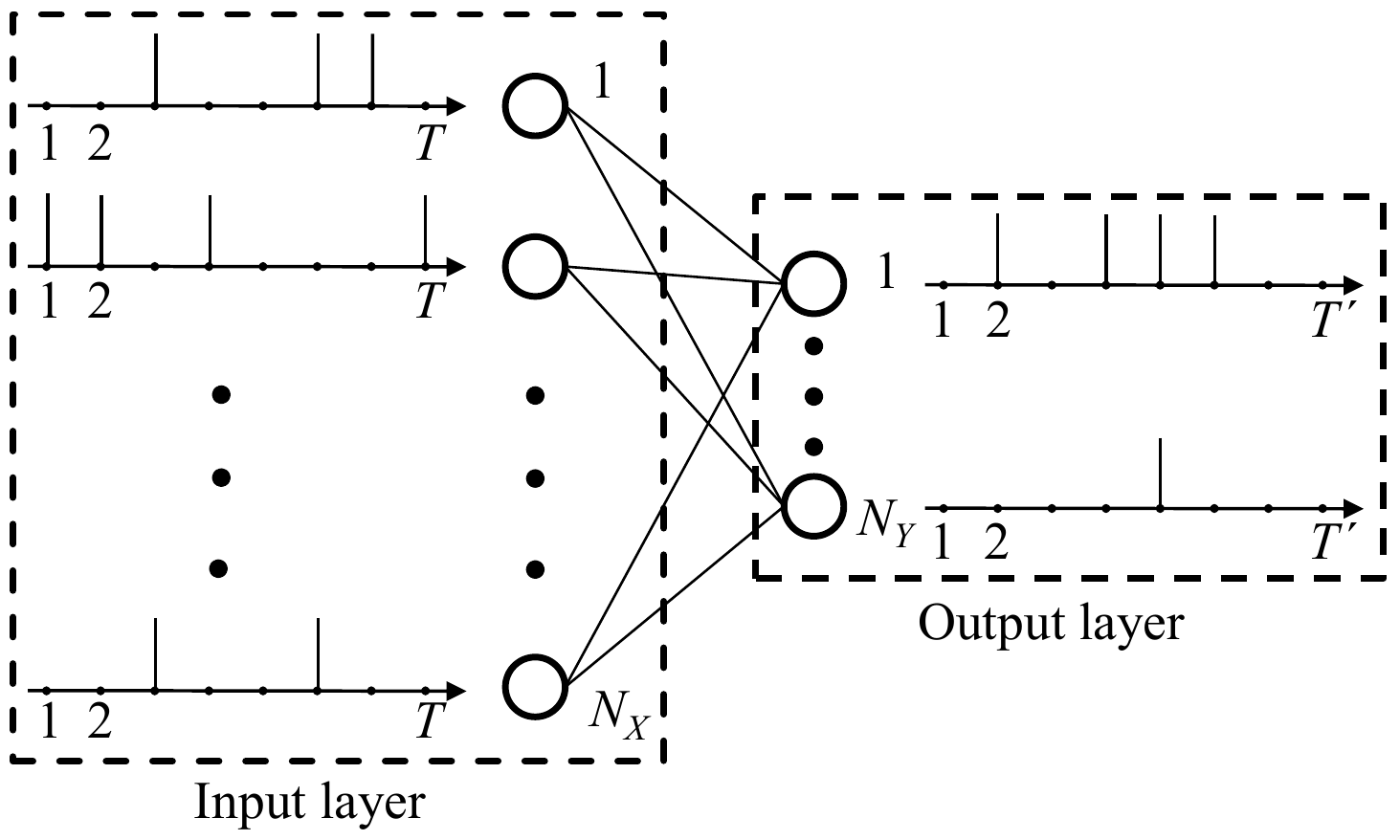}}
	\end{minipage}
	\caption{Two-layer SNN for supervised learning.}
	\label{fig:Net_diag}
	%
\end{figure}
%

While the use of probabilistic models for spiking neurons is standard in the context
of computational neuroscience (see, e.g., \cite{pillow2005prediction}), probabilistic modeling has been sparsely considered in the machine learning literature on SNNs. This is despite the known increased flexibility and expressive power of probabilistic models \cite{koller2009probabilistic, simeone2017brief}. In the context of SNNs, as an example, probabilistic models have the capability of learning firing thresholds using standard gradient based methods, while in deterministic models these are instead treated as hyperparameters and set by using heuristic mechanisms such as homeostasis \cite{jolivet2006predicting}. 
The state of the art on supervised learning with probabilistic models is set by \cite{gardner2016supervised} that considers Stochastic Gradient Descent (SGD) for Maximum Likelihood (ML) training, under the assumption that there exist given desired output spike trains for all output neurons.

In this paper, we study the problem of training the two-layer SNN illustrated in Fig. \ref{fig:Net_diag} under a probabilistic neuron model, for the purpose of classification. Conventional decoding in SNNs operates offline by selecting the output neuron, and hence the corresponding class, with the largest number of output spikes \cite{gardner2016supervised}. In contrast, we study here a first-to-spike decoding rule, whereby the SNN can perform an early classification decision once a spike firing is detected at an output neuron. This generally reduces decision latency and complexity during the inference phase. 
The first-to-spike decision method has been investigated with temporal, rather than rate, coding and deterministic neurons in \cite{masquelier2007unsupervised, mozafari2017first, wang2017spiketemp, lin2017relative}, but no learning algorithm exists under probabilistic neural models.

To fill this gap, we first propose the use of the flexible and computationally
tractable Generalized Linear Model (GLM) that was introduced in \cite{pillow2008spatio} in the context of computational neuroscience (Section \ref{sec:Conv_dec_sec}). Under this model, we then derive a novel SGD-based learning algorithm that maximizes the likelihood that the first spike is observed at the correct output neuron (Section \ref{sec:FSM_dec_sec}). Finally, we present numerical results that bring insights into the optimal parameter selection for the GLM neuron and on the accuracy-complexity trade-off performance of conventional and first-to-spike decoding rules.

%
\section {SPIKING NEURAL NETWORK WITH GLM NEURONS}\label{sec:SNN_Classification}
%
In this section, we describe the architecture of the two-layer SNN under study and then we present the proposed GLM neuron model.

%
\textbf{Architecture}
%
. We consider the problem of classification using a two-layer SNN. 
As shown in Fig. \ref{fig:Net_diag}, the SNN is fully connected and has $N_X$ presynaptic neurons in the input, or sensory layer, and $N_Y$ neurons in the output layer. Each output neuron is associated with a class. 
In order to feed the SNN, an input example, e.g., a gray scale image, is converted to a set of $N_X$ discrete-time spike trains, each with $T$ samples, through rate encoding. 
The input spike trains are fed to the $N_Y$ postsynaptic GLM neurons, which output discrete-time spike trains.
A decoder then selects the image class on the basis of the spike trains emitted by the output neurons. 

\textbf{Rate encoding}
%
. With the conventional rate encoding method, each entry of the input signal, e.g., each pixel for images, is converted into a discrete-time spike train by generating an independent and identically distributed (i.i.d.) Bernoulli vectors. The probability of generating a $``1"$, i.e., a spike, is proportional to the value of the entry. In the experiments in Sec. \ref{sec:Sim_sec}, we use gray scale images with pixel intensities between 0 and 255 that yield a spike probability between 0 and $1/2$.

\begin{figure}[t]
	\begin{minipage}[t]{1.0\linewidth}		
		\centering
		\centerline{\includegraphics[scale = 0.16]{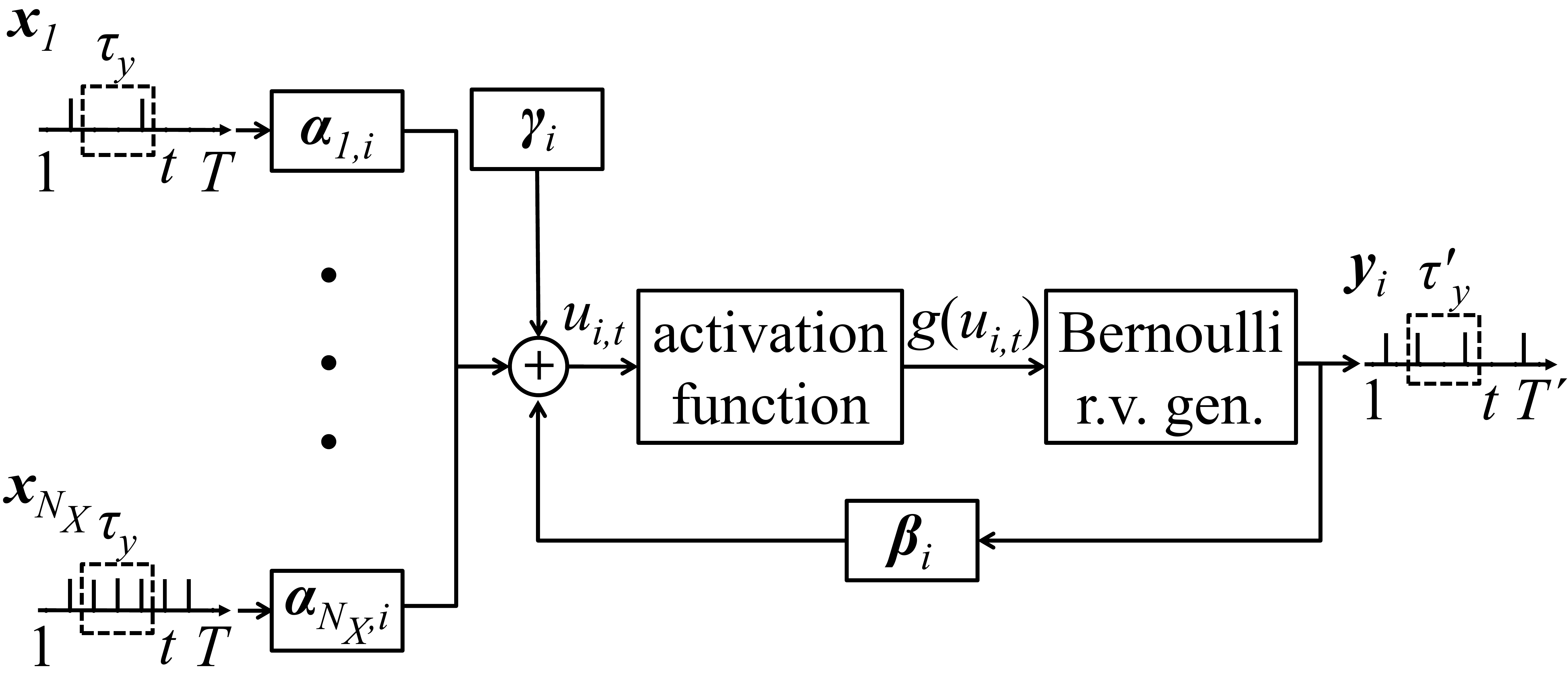}}
	\end{minipage}
	\caption{GLM neuron model.}
	\label{fig:GLM_diagram}
	%
\end{figure}

%
\textbf{GLM neuron model}
%
. The relationship between the input spike trains from the ${N_X}$ presynaptic neurons and the output spike train of any postsynaptic neuron $i$ follows a GLM, as illustrated in Fig. \ref{fig:GLM_diagram}.
To elaborate, we denote as ${x_{j,t}}$ and ${y_{i,t}}$ the binary signal emitted by the $j$-th presynaptic and the $i$-th postsynaptic neurons, respectively, at time $t$. Also, we let ${\bf{x}}_{j,a}^{b} = \left( {{x_{j,a}},...,{x_{j,b}}} \right)$
be the vector of samples from spiking process of the presynaptic neuron $j$ in the time interval $\left[ {a, b} \right]$. Similarly, the vector ${\bf{y}}_{i,a}^{b} = ({y_{i,a}},...,{y_{i,b}})$ contains samples from the spiking process of the neuron $i$ in the interval $\left[ {a, b} \right]$.
%
%
As seen in Fig. \ref{fig:GLM_diagram}, the output ${y_{i,t}}$ of postsynaptic neuron $i$ at time $t$ is Bernoulli distributed, with firing probability that depends on the past spiking behaviors $\{ {\bf{x}}_{j,t - {\tau _y}}^{t - 1}\} $ of the presynaptic neurons $j = 1,...,{N_X}$ in a window of duration $\tau_y$ samples, as well as on the past spike timings ${{\bf{y}}_{i,t - \tau _y^{\prime}}^{t - 1}} $ of neuron $i$ in a window of duration ${\tau_y^{\prime}}$ samples.
Mathematically, the membrane potential of postsynaptic neuron $i$ at time $t$ is given by
%
\begin{equation} \label{membrane_potential_GLM}
{u_{i,t}} = \sum\limits_{j = 1}^{{N_X}} {{\boldsymbol{\alpha }}_{j,i}^T{\bf{x}}_{j,t - {\tau _y}}^{t - 1}}  + {\boldsymbol{\beta }}_i^T{\bf{y}}_{i,t - {\tau_y^{\prime}}}^{t - 1} + {\gamma _i},
\end{equation}
%
where ${{\boldsymbol{\alpha }}_{j,i}} \in {\mathbb{R} ^{\tau_y} }$ is a vector that defines the \textit{synaptic kernel} (SK)
applied on the $\left\{ {j,i} \right\}$ synapse between presynaptic neuron  $j$ and postsynaptic neuron $i$; ${{\boldsymbol{\beta }}_i} \in {\mathbb{R}^{\tau _y^{\prime}}}$ is the \textit{feedback kernel} (FK); and ${{\gamma _i}}$ is a bias parameter.
The vector of variable parameters ${{\boldsymbol{\theta }}_i}$ includes the bias ${\gamma _i}$ and the parameters that define the SK and FK filters, which are discussed below.
Accordingly, the log-probability of the entire spike train ${{\bf{y}}_i} = {\left[ {{y_{i,1}},...,{y_{i,T}}} \right]^T}$ conditioned on the input spike trains ${\bf{x}} = \left\{ {{{\bf{x}}_j}} \right\}_{j = 1}^{{N_X}}$ can be written as
%
\begin{equation} \label{loglikelihood_BGLM}
\log {p_{{{\boldsymbol{\theta }}_i}}}\!\left( {{{\bf{y}}_i}\left| {\bf{x}} \right.} \right) = \sum\limits_{t = 1}^T {\left[ {{y_{i,t}}\log g\left( {{u_{i,t}}} \right) + {{\bar y}_{i,t}}\log \bar g\left( {{u_{i,t}}} \right)} \right]} ,
\end{equation}
where $g\left( \cdot \right)$ is an activation function, such as the sigmoid function $g\left( x \right) = \sigma \left( x \right) = 1/\left( {1 + \exp \left( { - x} \right)} \right)$, and we defined ${{\bar y}_{i,t}} = 1 - {y_{i,t}}$ and $\bar g\left( {{u_{i,t}}} \right) = 1 - g\left( {{u_{i,t}}} \right)$.

\begin{figure}[t]
	\begin{minipage}[t]{1.0\linewidth}
		\centering
		\centerline{\includegraphics[scale= 0.68]{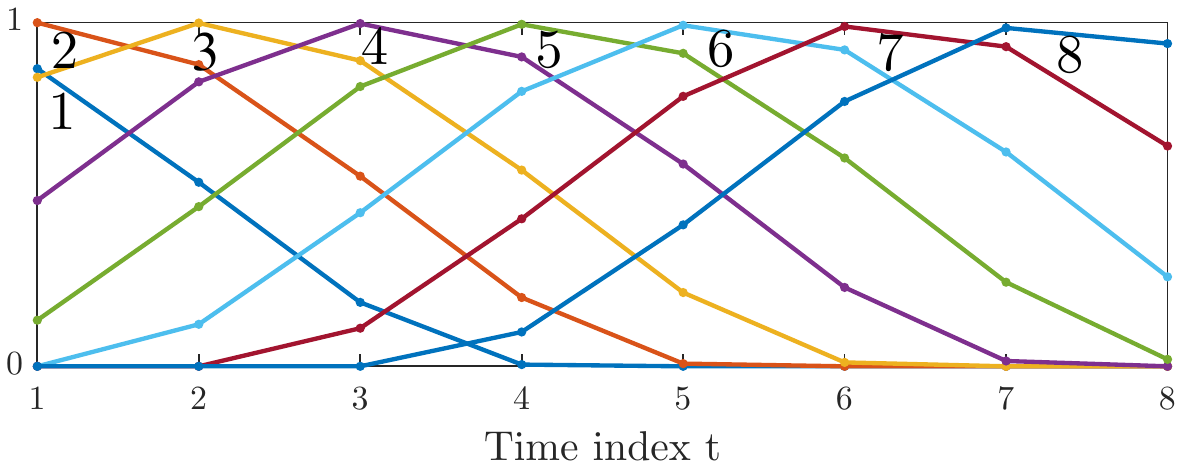}}
	\end{minipage}
	\caption{Basis functions used in Sec. \ref{sec:Sim_sec} ($a = 7500$ and $c = 1$ in \cite[Sec. Methods]{pillow2008spatio}).}
	\label{fig:Basis_funcs}
	%
\end{figure}

Unlike prior work on SNNs with GLM neurons, we adopt here the parameterized model introduced in \cite{pillow2008spatio} in the field of computational neuroscience. Accordingly, the SK and FK filters are parameterized as the sum of fixed basis functions with learnable weights. 
To elaborate, we write the SK ${{\boldsymbol{\alpha }}_{j,i}}$ and the FK ${{\boldsymbol{\beta }}_i}$ as
%
\begin{equation} \label{Ex_GLM_1}
{{\boldsymbol{\alpha }}_{j,i}} = {\bf{A}}{{\bf{w}}_{j,i}} ,
\text{ and }
{{\boldsymbol{\beta }}_{i}} = {\bf{B}}{{\bf{v}}_{i}} ,
\end{equation}
%
respectively, where we have defined the matrices ${\bf{A}} = \left[ {{{\bf{a}}_1},...,{{\bf{a}}_{{K_{\boldsymbol{\alpha }}}}}} \right]$ and ${\bf{B}} = \left[ {{{\bf{b}}_1},...,{{\bf{b}}_{{K_{\boldsymbol{\beta }}}}}} \right]$ and the vectors ${{\bf{w}}_{j,i}} = {\left[ {{w_{j,i,1}},...,{w_{j,i,{K_{\boldsymbol{\alpha }}}}}} \right]^T}$ and ${{\bf{v}}_i} = {\left[ {{v_{i,1}},...,{v_{i,{K_{\boldsymbol{\beta }}}}}} \right]^T}$; ${K_{\boldsymbol{\alpha }}}$ and ${K_{\boldsymbol{\beta }}}$ denote the respective number of basis functions; ${{\bf{a}}_k} = {\left[ {{a_{k,1}},...,{a_{k,{\tau _y}}}} \right]^T}$ and ${{\bf{b}}_k} = {\left[ {{b_{k,1}},...,{b_{k,{\tau _y^{\prime}}}}} \right]^T}$ are the basis vectors; and $\left\{ {{w_{j,i,k}}} \right\}$ and $\left\{ {{v_{i,k}}} \right\}$ are the learnable weights for the kernels ${{\boldsymbol{\alpha }}_{j,i}}$ and ${{\boldsymbol{\beta }}_i}$, respectively. 
This parameterization generalizes previously studied models for machine learning application. For instance, as a special case, if we set ${K_{\boldsymbol{\alpha }}} = {K_{\boldsymbol{\beta }}} = 1$, set ${{{\bf{a}}_1}}$ and ${{{\bf{b}}_1}}$ as in  \cite[eqs. (4) and (5)]{gardner2016supervised}, and fix the weights ${{v_{i,1} = 1}}$, equation \eqref{loglikelihood_BGLM} yields a discrete-time approximation of the model considered in \cite{gardner2016supervised}. As another example, if we set ${K_{\boldsymbol{\alpha }}} = \tau_y $, ${K_{\boldsymbol{\beta }}} = \tau _y^{\prime}$, ${{\bf{a}}_k} = {{\bf{1}}_k}$, ${{\bf{b}}_k} = {{\bf{1}}_k}$, where ${{\bf{1}}_k}$ is the all-zero vector except for a one in position $k$, \eqref{membrane_potential_GLM} yields the unstructured GLM model considered in \cite{shlens2014notes}.
For the experiments discussed in Sec. \ref{sec:Sim_sec}, we adopt the time-localized raised cosine basis functions introduced in \cite{pillow2008spatio}, which are illustrated in Fig. \ref{fig:Basis_funcs}.
Note that this model is flexible enough to include the learning of synaptic delays \cite{baldi1994delays, taherkhani2015dl}.
%

%
%
%
\section {TRAINING WITH CONVENTIONAL DECODING}\label{sec:Conv_dec_sec}
%
In this section, we briefly review ML training based on conventional rate decoding for the two-layer SNN. 
During the inference phase, decoding is conventionally carried out in post-processing by selecting the output neuron with the largest number of spikes. In order to facilitate the success of this decoding rule, in the training phase, the postsynaptic neuron corresponding to the correct label $c \in \left\{ {1,...,{N_Y}} \right\}$ is typically assigned a desired output spike train ${{\bf{y}}_c}$ with a number of spikes, while a zero output is assigned to the other postsynaptic neurons ${{\bf{y}}_i}$ with $i \ne c$. 

Using the ML criterion, one hence maximizes the sum of the log-probabilities \eqref{loglikelihood_BGLM} of the desired output spikes ${\bf{y}}\left( c \right) = \left\{ {{{\bf{y}}_1}\left( c \right),...,{{\bf{y}}_{{N_Y}}}\left( c \right)} \right\}$ for the correct label $c$ given the $N_X$ input spike trains ${\bf{x}} = \left\{ {{{\bf{x}}_1},...,{{\bf{x}}_{{N_X}}}} \right\}$, i.e.,
%
\begin{equation} \label{LL_conv}
{L}\left( {\boldsymbol{\theta }} \right) = \sum\limits_{i = 1}^{{N_Y}} {\log {p_{{{\bf{\boldsymbol{\theta} }}_i}}}\!\left( {\left. {{{\bf{y}}_i\left( c \right)}} \right|{\bf{x}}} \right)} .
\end{equation}
The sum is further extended to all examples in the training set. 
The parameter vector ${\boldsymbol{\theta }} = \left\{ {{\bf{W}},{\bf{V}},{\boldsymbol{\gamma }}} \right\}$ includes the parameters ${\bf{W}} = \left\{ {{{\bf{W}}_i}} \right\}_{i = 1}^{{N_Y}}$, ${\bf{V}} = \left\{ {{{\bf{v}}_i}} \right\}_{i = 1}^{{N_Y}}$ and ${\boldsymbol{\gamma }} = \left\{ {{\gamma _i}} \right\}_{i = 1}^{{N_Y}}$.
The negative log-likelihood $-{L}\left( {\boldsymbol{\theta }} \right)$ is convex with respect to ${\boldsymbol{\theta }}$ and can be minimized via SGD. 
For completeness, we report the gradients of ${L}\left( {\boldsymbol{\theta }} \right)$ in Appendix \ref{Appendix_A}.

%
\section {TRAINING WITH FIRST-TO-SPIKE DECODING}\label{sec:FSM_dec_sec}
%
In this section, we introduce the proposed learning approach based on GLM neurons and first-to-spike decoding.

During the inference phase, with first-to-spike decoding, a decision is made once a first spike is observed at an output neuron. In order to train the SNN for this classification rule, we propose to follow the ML criterion by maximizing the probability to have the first spike at the output neuron corresponding to the correct label $c$. The logarithm of this probability for a given example $\bf{x}$ can be written as 
%
\begin{equation} \label{first_spk_method_2}
L\left( {\boldsymbol{\theta }} \right) = \log \left( {\sum\limits_{t = 1}^T {{p_t}\left( {\boldsymbol{\theta }} \right)} } \right),
\end{equation}
where 
%
\begin{equation} \label{first_spk_method_3}
{p_t}\left( {\boldsymbol{\theta }} \right) = \prod\limits_{i = 1,i \ne c}^{{N_Y}} {\prod\limits_{t' = 1}^t {\bar g\left( {{u_{i,t'}}} \right)} } g\left( {{u_{c,t}}} \right)\prod\limits_{t' = 1}^{t - 1} {\bar g\left( {{u_{c,t'}}} \right)},
\end{equation}
is the probability of having the first spike at the correct neuron $c$ at time $t$. In \eqref{first_spk_method_3}, the potential $u_{i,t}$ for all $i$ is obtained from \eqref{membrane_potential_GLM} by setting $y_{i,t}  = 0$ for all $i$ and $t$.
The log-likelihood function $L\left( {\boldsymbol{\theta }} \right)$ in \eqref{first_spk_method_2} is not concave, and we tackle its maximization via SGD. 

To this end, the gradients of the log-likelihood function for a given input $\bf{x}$ can be computed after some algebra as (see Appendix \ref{Appendix_B} for details)
%
\begin{equation} \label{first_spk_method_w}
\begin{split}
&{\nabla _{{{\bf{w}}_{j,i}}}}L\left( {\boldsymbol{\theta }} \right) \\
&= \left\{ {\begin{array}{*{20}{l}}
	{ - \sum\limits_{t = 1}^T {{\rho _{i,t}}{h_t}g\left( {{u_{i,t}}} \right){{\bf{A}}^T}{\bf{x}}_{j,t - {\tau _y}}^{t - 1}} }&{i \ne c}\\
	{ - \sum\limits_{t = 1}^T {{\rho _{c,t}}\left( {{h_t}g\left( {{u_{c,t}}} \right) - {q_t}} \right){{\bf{A}}^T}{\bf{x}}_{j,t - {\tau _y}}^{t - 1}} }&{i = c}
	\end{array}} \right.,
\end{split}
\end{equation}
for the weights and
%
\begin{equation} \label{first_spk_method_gamma}
{\nabla _{{\gamma _i}}}L\left( {\boldsymbol{\theta }} \right) = \left\{ {\begin{array}{*{20}{l}}
	{ - \sum\limits_{t = 1}^T {{\rho _{i,t}}{h_t}g\left( {{u_{i,t}}} \right)} }&{i \ne c}\\
	{ - \sum\limits_{t = 1}^T {{\rho _{c,t}}\left( {{h_t}g\left( {{u_{c,t}}} \right) - {q_t}} \right)} }&{i = c}
	\end{array}} \right.,
\end{equation}
for the biases, where we have defined
%
%
\begin{equation} \label{first_spk_method_rho}
{\rho _{i,t}} = \frac{{g'\left( {{u_{i,t}}} \right)}}{{g\left( {{u_{i,t}}} \right)\bar g\left( {{u_{i,t}}} \right)}},
\end{equation}
and
%
%
\begin{equation} \label{first_spk_method_h}
	{h_{t}} = \sum\limits_{t' =  t}^T {{q_{t'}}}  = {1 - \sum\limits_{t' = {1}}^{t - 1} {{q_{t'}}} },
\end{equation}
with
%
\begin{equation} \label{first_spk_method_q}
{q_t} = \frac{{{p_t}\left( {\boldsymbol{\theta }} \right)}}{{\sum\limits_{t' = 1}^T {{p_{t'}}\left( {\boldsymbol{\theta }} \right)} }}.
\end{equation}
Note that we have ${\rho _{i,t}}  = 1$ when $g$ is the sigmoid function. 

Based on \eqref{first_spk_method_w}, the resulting SGD update can be considered as a neo-Hebbian rule \cite{fremaux2015neuromodulated}, since it multiplies the contributions of the presynaptic neurons and of the postsynaptic activity, where the former depends on $\bf{x}$ and the latter on the potential $u_{i,t}$.  
Furthermore, in \eqref{first_spk_method_w}${-}$\eqref{first_spk_method_gamma}, the probabilities $g\left( {{u_{i,t}}} \right)$ and $g\left( {{u_{c,t}}} \right)$ of firing at time $t$ are weighted by the probability $h_t$ in \eqref{first_spk_method_h}. By \eqref{first_spk_method_q}, this is the probability that the correct neuron is the first to spike and that it fires at some time $t' \ge t$, given that it is the first to spike at some time in the interval $\left[ {1,2,...,T} \right]$.

As a practical note, in order to avoid vanishing values in calculating the weights \eqref{first_spk_method_q}, we compute each probability term ${{p_t}\left( {\boldsymbol{\theta }} \right)}$ in the log-domain, and normalize all the resulting terms with respect to the minimum probability as ${q_t} = \exp \left( {{a_t}} \right)/\sum\nolimits_{t' = 1}^T {\exp \left( {{a_{t'}}} \right)} $, where ${a_t} = \ln \left( {{p_t}} \right) - {\min _t}\left( {\ln \left( {{p_t}} \right)} \right)$.

%
\begin{figure}[t]
	\begin{minipage}[b]{1.0\linewidth}
		\centering
		\centerline{\includegraphics[scale= 0.56]{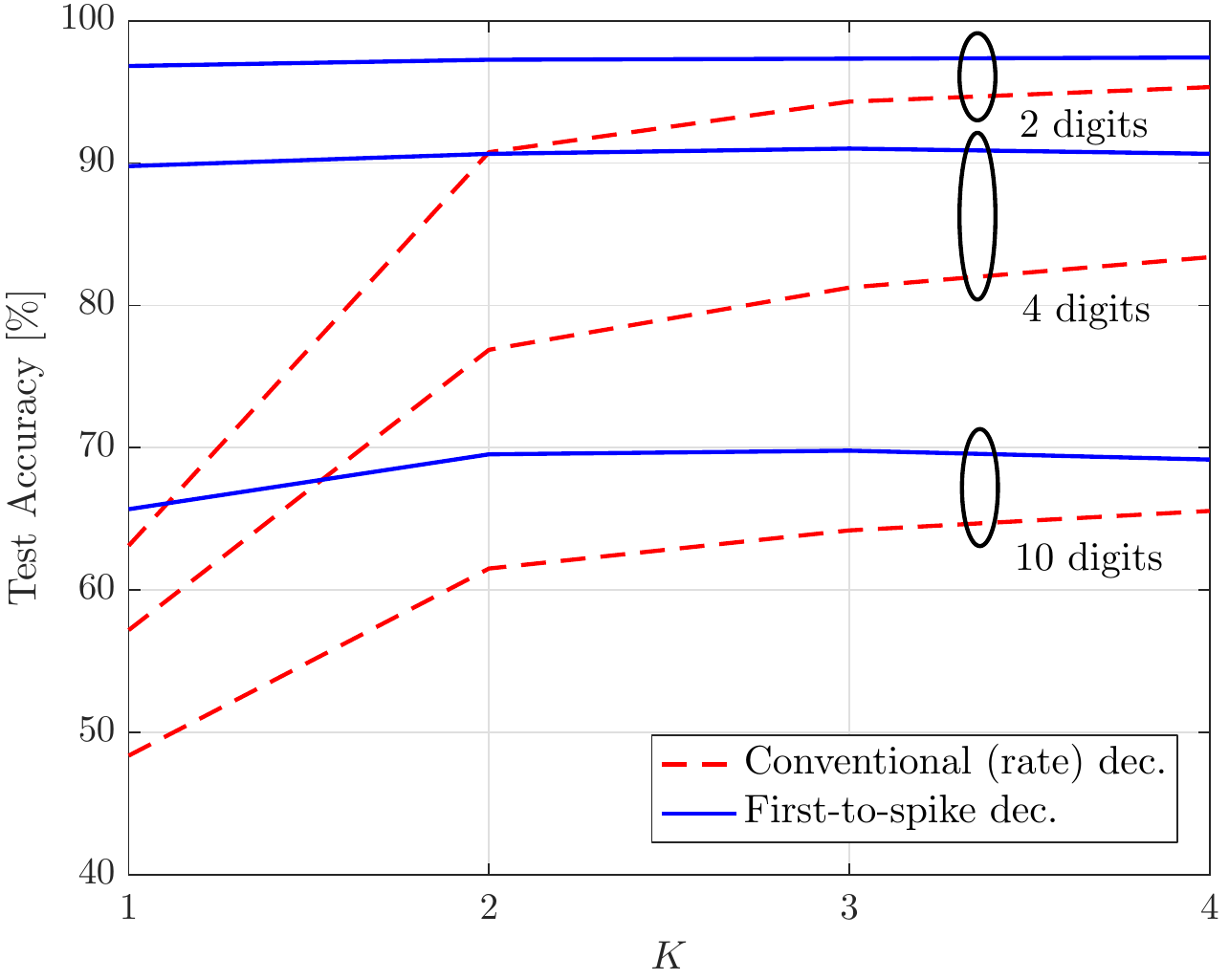}}
	\end{minipage}
	%
	{\caption{Test accuracy versus the number $K$ of basis functions for both conventional (rate) and first-to-spike decoding rules when $T = 4$.}}
	\label{fig:Dec_vs_K}
	%
\end{figure}
%
%

%
\section{NUMERICAL RESULTS}\label{sec:Sim_sec}
%
In this section, we numerically study the performance of the probabilistic SNN in Fig. \ref{fig:Net_diag} under conventional and first-to-spike decoding rules. We use the standard MNIST dataset \cite{lecun1998mnist} as the input data. As a result, we have $N_X = 784$, with one input neuron per pixel of the $\bf{x}$ images.
Following \cite{diamond2016comparing}, we consider different number of classes, or digits, namely, the two digits $\{5, 7\} $, the four digits $\{5, 7, 1, 9 \} $ and all 10 digits $\{0,...,9\}$, and we use 1000 samples of each class for training and the same number for test set.
We use a desired spike train with one spike after every three zeros for training the conventional decoding.
SGD with minibatch size of one with 200 training epochs is used for both schemes. 
Ten-fold cross-validation is applied for selecting between $10^{-3}$ or $10^{-4}$ for the constant learning rates. The model parameters $\boldsymbol{\theta}$ are randomly initialized with uniform distribution between -1 and 1. 

We evaluate the performance of the schemes in terms of classification accuracy in the test set and of inference complexity. The inference complexity is measured by the total number of elementary operations, namely additions and multiplications, for input image that are required by the SNN during inference. The number of arithmetic operations needed to calculate the membrane potential \eqref{membrane_potential_GLM} of neuron $i$ at time instant $t$ is of the order of $ {\cal O}\left( {{N_X}{\tau _y} + {{\tau _y^{\prime}}}} \right)$. As a result, in the conventional decoding method, the inference complexity per output neuron, or per class, is of the order ${\cal O}\left( {T\left( {{N_X}{s_{\bf{x}}} + {s_{{{\bf{y}}}}}} \right)} \right)$, where  ${s_{\bf{x}}}$ and ${s_{\bf{y}}}$ are the fraction of spikes in $\bf{x}$ and $\bf{y}$, respectively. In contrast, with the first-to-spike decoding rule, the SNN can perform an early decision once a single spike is detected, and hence its complexity order is ${\cal O}\left( {t\left( {{N_X}{s_{\bf{x}}} + {s_{{{\bf{y}}}}}} \right)} \right)$, where $1 \le t \le T$ is the (random) decision time.  

We first consider the test classification accuracy as a function of the number $K$ of basis functions in the GLM neural model. The basis functions are numbered as in Fig. \ref{fig:Basis_funcs}, and we set $T = 4$. 
From Fig. \ref{fig:Dec_vs_K}, we observe that conventional decoding requires a large number $K$ in order to obtain its best accuracy. This is due to the need to ensure that the correct output neuron fires consistently more than the other neurons in response to the input spikes. This, in turn, requires a larger temporal reception field, i.e., a larger $K$, to be sensitive to the randomly located input spikes. 
We note that for small values of $T$, such as $T = 4$, first-to-spike decoding obtains better accuracies than conventional decoding.

\begin{figure}[t]
	\begin{minipage}[b]{1.0\linewidth}
		\centering
		\centerline{\includegraphics[scale= 0.56]{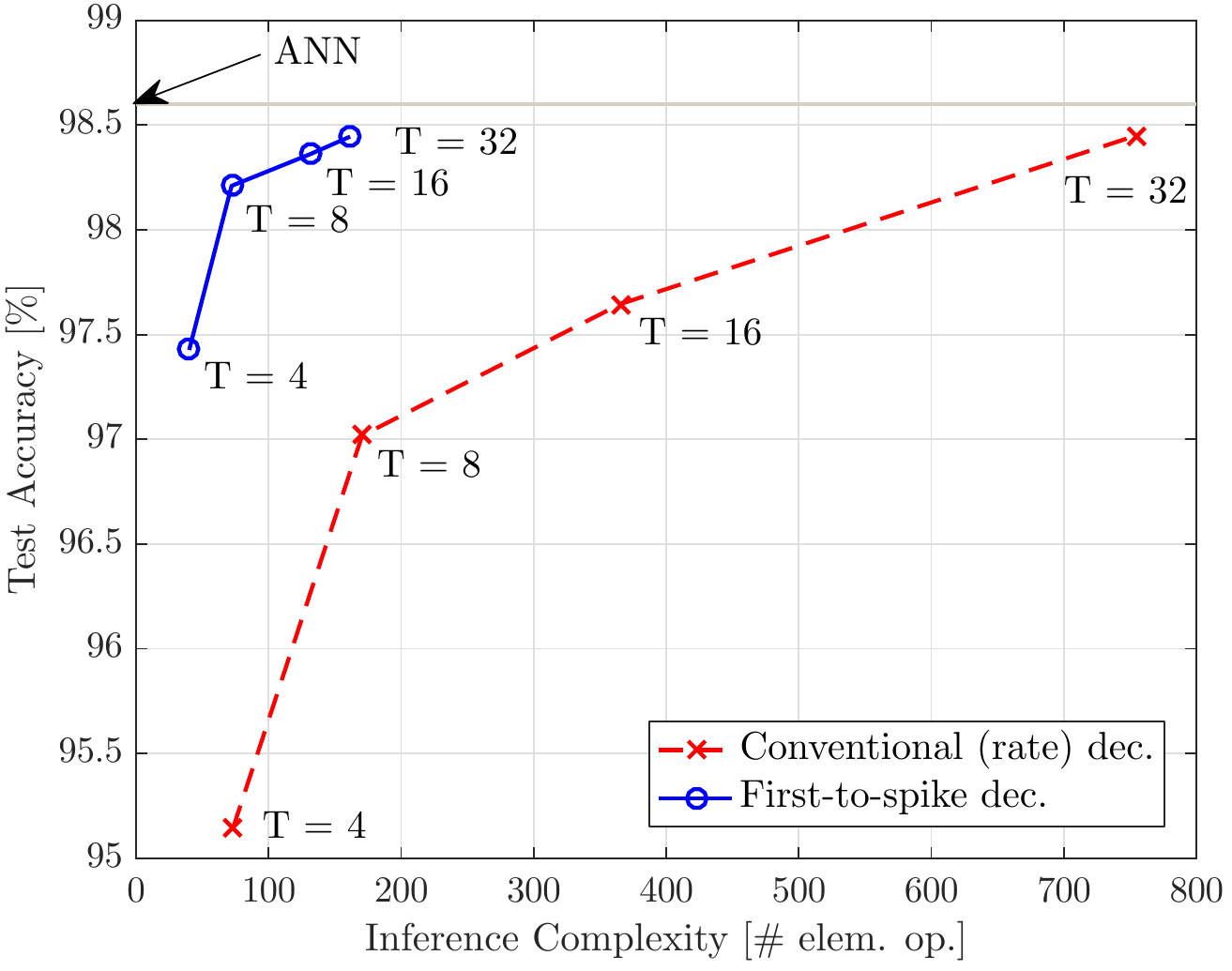}}
	\end{minipage}
	%
	\caption{Test accuracy versus per-class inference complexity for both conventional (rate) and first-to-spike decoding rules.}
	\label{fig:Dec_vs_Comp}
	%
\end{figure}
%

Fig. \ref{fig:Dec_vs_Comp} depicts the test classification accuracy versus the inference complexity for both conventional and first-to-spike decoding rules for two digits when $K = T$. The classification accuracy of a conventional two-layer artificial neural network (ANN) with logistic neurons is added for comparison. From the figure, first-to-spike decoding is seen to offer a significantly lower inference complexity, thanks to its capability for early decisions, without compromising the accuracy. 
For instance, when the classification accuracy equals to $98.4\%$, the complexity of the conventional decoding method is five times larger than the first-to-spike method.
Note also that conventional decoding generally requires large values of $T$ to perform satisfactorily.  

%
\section{CONCLUSIONS}\label{sec:Conclusion_sec}
%
In this paper, we have proposed a novel learning method for probabilistic two-layer SNN that operates according to the first-to-spike learning rule. We have demonstrated that the proposed method improves the accuracy-inference complexity trade-off with respect to conventional decoding. Additional work is needed in order to generalize the results to multi-layer networks.


\section{ACKNOWLEDGMENT}
This work was supported by the U.S. NSF under grant ECCS $\#$1710009. O. Simeone has also received funding from the European Research Council (ERC) under the European Union’s Horizon 2020 research and innovation program (grant agreement $\#$725731).

\begin{center}
	
	\section*{{APPENDIX}}
	\pdfbookmark[section]{APPENDIX}{Appendix}
\end{center}
%
\subsection{{GRADIENTS FOR CONVENTIONAL DECODING}}\label{Appendix_A}
%

For a given input $\bf{x}$, the gradients of the log-likelihood function $L\left( {\boldsymbol{\theta }} \right)$ in \eqref{LL_conv} for conventional decoding  are given as
\begin{equation} \label{w_update_g}
{\nabla _{{{\boldsymbol{w}}_{j,i}}}}{L}\left( {\boldsymbol{\theta }} \right) = \sum\limits_{t = 1}^T {{e_{i,t}}{\rho _{i,t}}{{\bf{A}}^T}{\bf{x}}_{j,t - {\tau _y}}^{t - 1}}   ,
\end{equation}
\begin{equation} \label{v_update_g}
{\nabla _{{{\boldsymbol{v}}_i}}}{L}\left( {\boldsymbol{\theta }} \right) = \sum\limits_{t = 1}^T {{e_{i,t}}{\rho _{i,t}}{{\bf{B}}^T}{\bf{y}}_{i,t - {\tau _y^{\prime}}}^{t - 1}}  ,
\end{equation}
and
\begin{equation} \label{gamma_update_g}
{\nabla _{{\gamma _i}}}{L}\left( {\boldsymbol{\theta }} \right) = \sum\limits_{t = 1}^T {{e_{i,t}}{\rho _{i,t}}} ,
\end{equation}
where 
\begin{equation} \label{err}
{e_{i,t}} = {y_{i,t}} - g\left( {{u_{i,t}}} \right),
\end{equation}
is the error signal, and ${\rho _{i,t}}$ is given as
\begin{equation} \label{rho_def}
{\rho _{i,t}}  = \frac{{g'\left( {{u_{i,t}}} \right)}}{{g\left( {{u_{i,t}}} \right)\bar g\left( {{u_{i,t}}} \right)}},
\end{equation}
where $g'\left( {{u_{i,t}}} \right) \buildrel \Delta \over = \frac{{dg\left( {{u_{i,t}}} \right)}}{{d{u_{i,t}}}}$.

\subsection{{CALCULATION OF GRADIENTS FOR FIRST-TO-SPIKE DECODING}}\label{Appendix_B}
The gradient of $L\left( {\boldsymbol{\theta }} \right)$ with respect to ${{{\bf{w}}_{j,i}}}$ for ${i \ne c}$ can be calculated as:
\begin{equation} \label{L_w_i_1}
\begin{split}
{\nabla _{{{\bf{w}}_{j,i}}}}L\left( {\boldsymbol{\theta }} \right) &= {\nabla _{{{\bf{w}}_{j,i}}}}\log \left( {\sum\limits_{t = 1}^T {{p_t}\left( {\boldsymbol{\theta }} \right)} } \right)\\
&= \frac{{\sum\limits_{t = 1}^T {{\nabla _{{{\bf{w}}_{j,i}}}}{p_t}\left( {\boldsymbol{\theta }} \right)} }}{{\sum\limits_{t = 1}^T {{p_t}\left( {\boldsymbol{\theta }} \right)} }}\\
&= \sum\limits_{t = 1}^T {{k_{ i,t}}{\nabla _{{{\bf{w}}_{j,i}}}}\prod\limits_{t' = 1}^t {\bar g\left( {{u_{i,t'}}} \right)} } ,
\end{split}
\end{equation}
where
\begin{equation} \label{K_w_i_t}
{k_{ i,t}} = \frac{{\prod\limits_{i' = 1,i' \ne i,c}^{{N_Y}} {\prod\limits_{t' = 1}^t {\bar g\left( {{u_{i',t'}}} \right)} } g\left( {{u_{c,t}}} \right)\prod\limits_{t' = 1}^{t - 1} {\bar g\left( {{u_{c,t'}}} \right)} }}{{\sum\limits_{t' = 1}^T {{p_{t'}}\left( {\boldsymbol{\theta }} \right)} }}.
\end{equation}

Using the generalized product rule for derivative of $k$ factors \cite{wiki} as 
\begin{equation} \label{L_w_i_2}
\begin{split}
\frac{d}{{dx}}\prod\limits_{i = 1}^k {{f_i}\left( x \right)}  &= \sum\limits_{i = 1}^k {\left( {\frac{d}{{dx}}{f_i}\left( x \right)\prod\limits_{j \ne i} {{f_j}\left( x \right)} } \right)} \\
& = \left( {\prod\limits_{i = 1}^k {{f_i}\left( x \right)} } \right)\left( {\sum\limits_{i = 1}^k {\frac{{{{f'}_i}\left( x \right)}}{{{f_i}\left( x \right)}}} } \right) ,
\end{split}
\end{equation}
we have
\begin{equation} \label{L_w_i_3}
\begin{split}
&{\nabla _{{{\bf{w}}_{j,i}}}}\prod\limits_{t' = 1}^t {\bar g\left( {{u_{i,t'}}} \right)}  =  - \prod\limits_{t' = 1}^t {\bar g\left( {{u_{i,t'}}} \right)} \sum\limits_{t' = 1}^t {\frac{{g'\left( {{u_{i,t'}}} \right)}}{{\bar g\left( {{u_{i,t'}}} \right)}}{\nabla _{{{\bf{w}}_{j,i}}}}{u_{i,t'}}} \\
&=  - \prod\limits_{t' = 1}^t {\bar g\left( {{u_{i,t'}}} \right)} \sum\limits_{t' = 1}^t {{\rho _{i,t'}}g\left( {{u_{i,t'}}} \right){\nabla _{{{\bf{w}}_{j,i}}}}{u_{i,t'}}},
\end{split}
\end{equation}
where we have used the equality $\bar g'\left( u \right) =  - g'\left( u \right)$ and ${\rho _{i,t}}$ is defined as in \eqref{first_spk_method_rho}. After substituting \eqref{L_w_i_3} into \eqref{L_w_i_1}, we have
\begin{equation} \label{L_w_i_4}
\begin{split}
{\nabla _{{{\bf{w}}_{j,i}}}}L\left( {\boldsymbol{\theta }} \right) & = - \sum\limits_{t = 1}^T {{q_t}\sum\limits_{t' = 1}^t {{\rho _{i,t'}}g\left( {{u_{i,t'}}} \right){\nabla _{{{\bf{w}}_{j,i}}}}{u_{i,t'}}} } \\
& = { - \sum\limits_{t = 1}^T {{h_t}{\rho _{i,t}}g\left( {{u_{i,t}}} \right){\nabla _{{{\bf{w}}_{j,i}}}}{u_{i,t}}} } ,
\end{split} 
\end{equation}
where we have defined ${q_t} $ and ${h_t} $ as in \eqref{first_spk_method_q} and \eqref{first_spk_method_h}, respectively. Given that we have the equality ${{\nabla _{{{\bf{w}}_{j,i}}}}{u_{i,t}}} = {{{\bf{A}}^T}{\bf{x}}_{j,t - {\tau _y}}^{t - 1}}$, we have \eqref{first_spk_method_w} for ${\nabla _{{{\bf{w}}_{j,i}}}}L\left( {\boldsymbol{\theta }} \right)$ when ${i \ne c}$. 

The gradient of $L\left( {\boldsymbol{\theta }} \right)$ with respect to ${{{\bf{w}}_{j,i}}}$ for ${i = c}$ can be calculated as:
\begin{equation} \label{L_w_c_1}
\begin{split}
{\nabla _{{{\bf{w}}_{j,c}}}}L\left( {\boldsymbol{\theta }} \right) = \sum\limits_{t = 1}^T {{k_{c,t}}{\nabla _{{{\bf{w}}_{j,c}}}}\left( {g\left( {{u_{c,t}}} \right)\prod\limits_{t' = 1}^{t - 1} {\bar g\left( {{u_{c,t'}}} \right)} } \right)} ,
\end{split}
\end{equation}
where
\begin{equation} \label{K_w_c_1}
{k_{c,t}} = {\frac{{\prod\limits_{i = 1,i \ne c}^{{N_Y}} {\prod\limits_{t' = 1}^t {\bar g\left( {{u_{i,t'}}} \right)} } }}{{\sum\limits_{t' = 1}^T {{p_{t'}}\left( {\boldsymbol{\theta }} \right)} }}} .
\end{equation}
Using \eqref{L_w_i_2}, we have
\begin{equation} \label{L_w_c_2}
\begin{split}
&{\nabla _{{{\bf{w}}_{j,c}}}}L\left( {\boldsymbol{\theta }} \right)\\ &= \sum\limits_{t = 1}^T {{q_t}\left[ {\frac{{g'\left( {{u_{c,t}}} \right)}}{{g\left( {{u_{c,t}}} \right)}}{\nabla _{{{\bf{w}}_{j,c}}}}{u_{c,t}} - \sum\limits_{t' = 1}^{t - 1} {\frac{{g'\left( {{u_{c,t'}}} \right)}}{{\bar g\left( {{u_{c,t'}}} \right)}}{\nabla _{{{\bf{w}}_{j,c}}}}{u_{c,t'}}} } \right]}  \\
& = \sum\limits_{t = 1}^T {{q_t}\left[ {{\rho _{c,t}}{\nabla _{{{\bf{w}}_{j,c}}}}{u_{c,t}} - \sum\limits_{t' = 1}^t {{\rho _{c,t'}}g\left( {{u_{c,t'}}} \right){\nabla _{{{\bf{w}}_{j,c}}}}{u_{c,t'}}} } \right]} .
\end{split}
\end{equation}
Thus, by substituting ${{\nabla _{{{\bf{w}}_{j,c}}}}{u_{c,t}}} = {{{\bf{A}}^T}{\bf{x}}_{j,t - {\tau _y}}^{t - 1}}$ into \eqref{L_w_c_2} and after simplification, equation \eqref{first_spk_method_w} is obtained for ${i = c}$, which completes the proof. Note that, the same procedure is done for ${\nabla _{{{\gamma}_{i}}}}L\left( {\boldsymbol{\theta }} \right)$ by considering the equality ${{\nabla _{{{\gamma}_{i}}}}{u_{i,t}}} = 1$ for all $i$.



\bibliographystyle{IEEEtran}
\bibliography{refs}
\end{document}